\documentclass[10pt,twocolumn,letterpaper]{article}

\usepackage{iccv}
\usepackage{times}
\usepackage{epsfig}
\usepackage{graphicx}
\usepackage{amsmath}
\usepackage{amssymb}
\usepackage{booktabs}
\usepackage{multirow}
\usepackage{caption}
\usepackage{subcaption}
\usepackage{authblk}

\usepackage[pagebackref=true,breaklinks=true,letterpaper=true,colorlinks,bookmarks=false]{hyperref}

\iccvfinalcopy 

\def\iccvPaperID{****} 

\ificcvfinal\fi

\begin{document}

\title{InAugment: Improving Classifiers via Internal Augmentation}

\author[1]{Moab Arar}
\author[2]{Ariel Shamir}
\author[1]{Amit Bermano}
\affil[1]{Tel-Aviv University}
\affil[2]{The Interdisciplinary Center Herzliya}

\maketitle
\ificcvfinal\thispagestyle{empty}\fi

\begin{abstract}
Image augmentation techniques apply transformation functions such as rotation, shearing, or color distortion on an input image. These augmentations were proven useful in improving neural networks' generalization ability. In this paper, we present a novel augmentation operation, InAugment, that exploits image internal statistics. The key idea is to copy patches from the image itself, apply augmentation operations on them, and paste them back at random positions on the same image. This method is simple and easy to implement and can be incorporated with existing augmentation techniques. We test InAugment on two popular datasets -- CIFAR and ImageNet. We show improvement over state-of-the-art augmentation techniques. Incorporating InAugment with Auto Augment yields a significant improvement over other augmentation techniques (e.g., +1\% improvement over multiple architectures trained on the CIFAR dataset). We also demonstrate an increase for ResNet50 and EfficientNet-B3 top-1's accuracy on the ImageNet dataset compared to prior augmentation methods. Finally, our experiments suggest that training convolutional neural network using InAugment not only improves the model's accuracy and confidence but its performance on out-of-distribution images. \footnote{The code will be made publicly available at \url{https://github.com/moabarar/inaugment} }
\end{abstract}

\section{Introduction}
Data augmentation is a popular technique that generates new instances based on some processing of available training data to increase its amount and variance. For image classification tasks, data augmentation is useful in improving the network's generalization, performance, and robustness. This improvement, for example, promotes invariance to fundamental transformations, such as scale, rotation, or color changes. 

Some augmentation techniques use handcrafted operations (e.g., CutOut~\cite{CutOut}), while others learn the desired operation needed to achieve the most accurate results (e.g., AutoAugment~\cite{AutoAugment})).  Still, most augmentation techniques create new images using some global operation (e.g., spatial transformation) or combine images to increase the input space variance~\cite{CutMix, Mixup, ManifoldMixup}. Despite the improvements achieved by these techniques, the final augmented image maintains a single view of the input, limiting each image's information variance.  This limitation can hurt the model's generalization ability and robustness, especially for out-of-distribution images.

Our work draws inspiration from the line of works showing that natural images have unique internal statistics: small patches of the image recur abundantly within itself~\cite{Inner2, Inner4}. Exploiting this property, we show that repeating patches at different scales help neural networks model the image's internal distribution, creating an image-specific prior. This property was utilized to solve many ill-posed vision tasks in an unsupervised manner~\cite{Inner1,Inner2, Inner3}. Nonetheless, to the best of our knowledge, previous data augmentation methods do not exploit this property.

\begin{figure*}[ht]
    \centering
    \includegraphics[width=\textwidth]{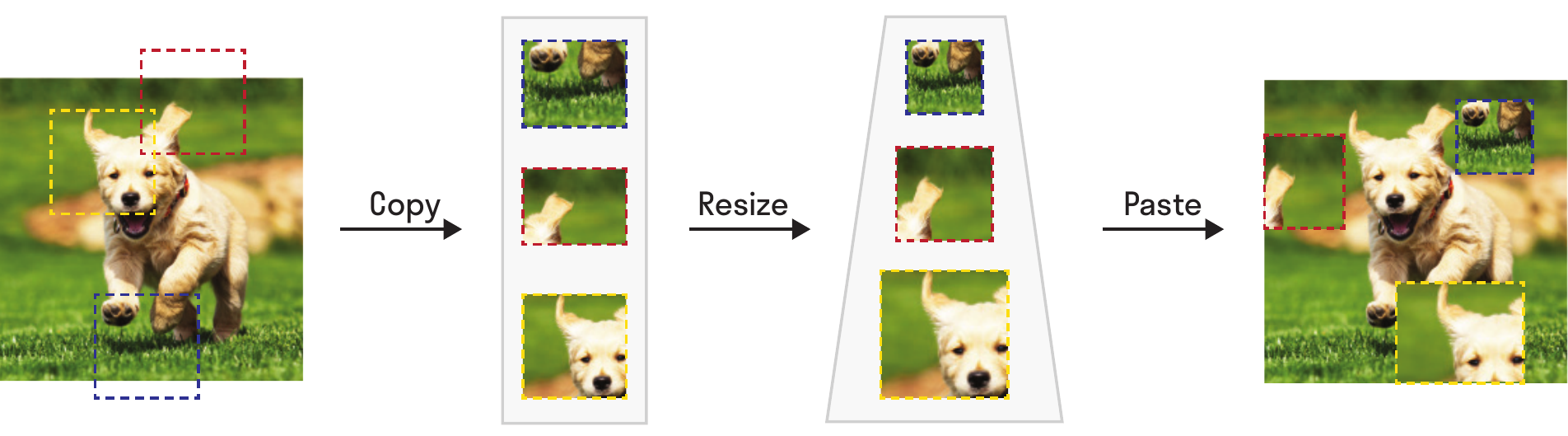}
    \caption{\textbf{InAugment overview.} We randomly extract patches from the input image (left). Each patch is then rescaled, and together they form a multi-scale resolution of the same input-image (middle). Finally, the multi-scale patches are pasted on the input image (right) to create the final augmented image. Note that since larger patches are more likely to occlude smaller ones, we sort the paste order according to the patch size.}
    \label{fig:overview}
\end{figure*}

In this paper, we present a novel data augmentation technique we call \emph{InAugment}. In the spirit of the aforementioned works, we seek to increase the information variance within each image rather than the entire dataset. This variance should be increased across different scales, as this is the behavior that natural images present. To adhere to these principles, we introduce a simple to implement two-step augmentation scheme, which can be incorporated in addition to other augmentation methods. 
In the first stage, we copy random patches from the input image. The patches are then resized, promoting multi-resolution views of the information in the image. In the second stage, the patches are randomly pasted back into the \textbf{same image}. This two-stage operation ensures that the image internal-information is repeated at different scales, helping neural networks capture the input image's inner-distribution prior. This patch repetition is beneficial when the patches that contain specific information about the target class (e.g., a dog's head) are pasted on non-salient regions (e.g., the background). In this case, the target class's main features are repeated in different scales,  encouraging the network to detect the input image's essential features.  

We validate our approach on the CIFAR~\cite{CIFAR} and ImageNet~\cite{ILSVRC12} datasets through training various standard bench-marking architectures using InAugment and several baselines. Throughout the experiments, we witness consistent improvements in the networks' accuracy, which are even more pronounced when we paste multiple patches at different scales. In particular, we gain between 1\% to 1.5\%  improvement over state-of-the-art augmentation techniques. We further show that models trained with InAugment, are more robust and handle out-of-distribution images really well. Our experiments demonstrate that InAugment should be added as an additional augmentation step when designing new augmentation approaches.
\section{Related work}
The ability of augmentation methods to improve deep models' generalization led to extensive research in this field. Basic operations such as random horizontal flip and crop became standard in classification tasks~\cite{AlexNet, VGG, ResNet, WideResNet}, while other techniques were proven to be useful for specific datasets. For example, approaches that learn spatial distortion~\cite{MnistSpatial1, MnistSpatial2, MnistSpatial3,MnistSpatial4}, are helpful on the MNIST~\cite{MNIST} dataset. For the CIFAR~\cite{CIFAR} and SVHN~\cite{CIFAR} datasets, region drop-out in the input forces the network to look at other features in the image. Unlike these methods, we show that InAugment achieves improvement on both small and large scale datasets.

Increasing the data can be achieved by mixing input instances. MixUp~\cite{Mixup} achieves impressive results via linear interpolation of the two input images while expecting the class labels to follow the same convex combination. Following MixUP, other mixing strategies showed improvements in the robustness~\cite{ManifoldMixup} and localization~\cite{CutMix} of neural networks. These methods operate in batch-resolution, which means data points are combined after the preprocessing stage. Specifically, since we perform InAugment in the preprocessing step, it can be added to existing mixing augmentations~\cite{Mixup, CutMix}.

Generative methods can also be used to increase the data size. Lemley et al.~\cite{SmartAug} train a neural network that incorporates images from the same class to reduce the classification loss. In~\cite{Baysian}, Tran et al. simultaneously train a classification network and a Bayesian network that generates augmented data, which help improve the classifier performance. Lastly, generative adversarial methods~\cite{GAN0} demonstrate excellent ability to generate high-quality images~\cite{StyleGAN}, and are leveraged for augmenting data by generating synthetic data~\cite{GAN1, GAN2, GAN3, GAN4} or domain-specific image transformations~\cite{GANForLearningTransformation}. These methods are usually used in the case where acquiring data is difficult (e.g, medical imaging), or for specific domains (e.g., human faces~\cite{StyleGAN}), but they don't perform well on data with high variance of natural images.

The complexity of designing hand-crafted augmentation methods gave rise to the automated search for augmentation policies. Recently, a novel method named AutoAugment~\cite{AutoAugment} uses reinforcement learning to search for a given dataset's best augmentation policy. The augmentation policy found consists of several sub-policies, and each sub-policy is composed of two basic image transformation operations (e.g., Rotation, Shearing, or Color Jitter). For each image, a sub-policy is chosen uniformly at random and is applied to produce the augmented image. One drawback of AutoAugment is the vast search space, which requires extensive computation power. Several methods, such as Fast-AutoAugment~\cite{FastAutoAugment}, Population-Based-Augmentation~\cite{PBA}, RandAugment~\cite{RandAugment} and Adversarial AutoAugment~\cite{AdverserialAutoAugment} try to reduce the search complexity while maintaining competitive results. Additionally, in a recent paper~\cite{OutliersAutoAugment}, Wei et al. argue that automated augmentation policies could lead to over distortion in the data. To overcome this, in addition to ground-truth labels, they use soft-labels provided by a teacher model to compensate for any semantic loss. 

In our work, we employ AutoAugment~\cite{AutoAugment} as the baseline augmentation, and show that InAugment boosts its performance; this suggests that policies found by other methods~\cite{FastAutoAugment, RandAugment, PBA} could also benefit from our method.

\section{Method}
Our method consists of two stages, a \textit{copy-stage},and a \textit{paste-stage}. In the copy-stage, we copy patches randomly from the underlying image, resize them and apply a base augmentation on the patches. The same base-augmentation is also performed on the input image. Later, during the paste-stage, we paste the augmented patches one after the other, according to their sizes, and produce the final augmented output (see figure~\ref{fig:overview}). In the following subsections, we give an in-depth description about different parts of InAugment. We begin by giving a brief overview of the base-augmentation we use, which is based on AutoAugment~\cite{AutoAugment}. Then we describe in detail the copy-and-paste stages. Throughout this section we let $I \in \mathbb{R}^{H\times W \times 3}$ be an input image of height $H$ and width $W$. 

\subsection{Base augmentation} 
\label{sec:base_augmentation}
We define an augmentation sub-policy to be an ordered set of transformations $T = \{o_1, o_2, \dots, o_k\}$, such that when $T$ is applied on an image $I$ the resulting image is:

\begin{equation}
T(I) = o_1 \circ o_2 \dots{} \circ o_k (I)
\end{equation}

Each operation $o_i$ represents a basic image transformation function (e.g, rotation), and is performed on the image with some predefined probability. The operation $\circ$ denotes the composition operation. 

In our implementation, we consider sub-policies used in the AutoAugment~\cite{AutoAugment} method. These sub-policies consists of two transformations (i.e., $k=2$). The operations used in AutoAugment are: \textit{Shear-x/y, Translate-x/y, Rotate, AutoContrast,  Invert,  Equalize,  Solarize,  Posterize,  Contrast,  Color,  Brightness, and  Sharpness}. Note, AutoAugment also uses CutOut~\cite{CutOut}, but we omit it from our implementation as our method operates as a region-drop as well (see subsection~\ref{sec:InaugmentVsCutout}). For further information about these sub-policies, please refer to~\cite{AutoAugment}. 

The final augmentation is determined by a policy, which is a finite set of sub-policies, i.e., $\mathcal{T} = \{ T_1, T_2, \dots{}\}$. To generate the augmented image, we randomly draw a sub-policy $T_i \in \mathcal{T}$, and apply it on $I$. 



\subsection{Copy stage}
\label{sec:copy_stage}
In the copy-stage, we randomly copy $n$-patches of size $H_p \times W_p$. The patch size can be randomly chosen (for each patch), or it can be set to some fixed value. In our implementation, we choose fixed-size patches for low-resolution images, and random-size patches when the training set has varying image sizes. In the case where the patches deviate from the image boundary, then we trim the patch to be only inside the image (this means that we don't use padding, which is commonly used for random copy). We denote $P=\{P_1, P_2, \dots, P_n\}$ to be the set of copied patches. Once the patches are copied, we perform resizing and augmentation on them. We consider two implementations for the following stage, namely, \textit{resize-first} and \textit{augment-first}.

\textbf{Resize-first:} in the resize first implementation, we first resize the copied patches, and only then apply the base augmentation on them. Formally, given an ordered set of target patch sizes $S_1, S_2, \dots S_n$, we let  $\tt{Resize}\left(P_i, S_i\right)$ be the operation that resizes the patch $P_i$ to the target size $S_i$. After we resize the patches, the augmentation $\mathcal{T}$ is applied, yielding the final copied patch for this implementation to be:

\begin{equation}
P_{copied} = \{ \mathcal{T}\left(\tt{Resize}\left(P_i,S_i\right)\right)\}_{i=1}^{n} 
\end{equation}

\textbf{Augment-first:} in this implementation, as the name suggests, we first apply the base augmentation and then resize the augmented patch. Therefore, the final copied patches in this implementation are:
\begin{equation}
P_{copied} = \{ \tt{Resize}\left(\mathcal{T}\left(P_i\right),S_i\right)\}_{i=1}^{n}    
\end{equation}

In our implementation, we found that for small images (e.g., images for the CIFAR~\cite{CIFAR} dataset), it is best to use \textbf{Augment-first} implementation. For high-resolution images (e.g., images from the ImageNet~\cite{ILSVRC12} dataset), we use the Resize-first implementation since it is more efficient to perform the base augmentation on smaller patches (after the resize). Furthermore, we found that it is better to apply the same transformations on the input image and the copied patches. We believe that sampling different augmentations for different patches will most likely yield a patch augmented with an easier transformation than the other patches. This, in turn, will bias the network to concentrate on easier patches. Note that we omit to transform the base-image first for efficiency reasons. It is better to copy smaller patches and perform the augmentation on smaller images than having to perform the augmentation on the entire image.

\subsection{Paste stage}
\label{sec:paste_stage}
Let $P_{copied}$ be the copied patches after resize and augmentation from the previous stage. Then, we paste the patches at random locations on top of the image $\mathcal{T}\left(I\right)$ (i.e, the image after the augmentation). We also drop patches with probability $1-p_i$, meaning, each patch in $P_{copied}$ is pasted onto the image with probability $p_i$. Note, the paste order follows the patches' sizes in $P_{copied}$. This means that we first paste the largest patch, and finally, we paste the smallest patch. This ensures that larger patches will not occlude smaller ones. Furthermore, to make the implementation as simple as possible, in the case where the pasted patch deviates from the image boundary, we clip it to fit the final augmented image.

\begin{figure}
     \centering
     \begin{subfigure}[b]{0.45\textwidth}
         \centering
         \includegraphics[width=\textwidth]{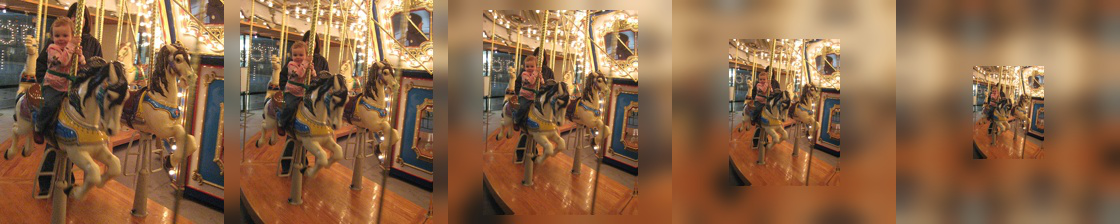}
         \includegraphics[width=\textwidth]{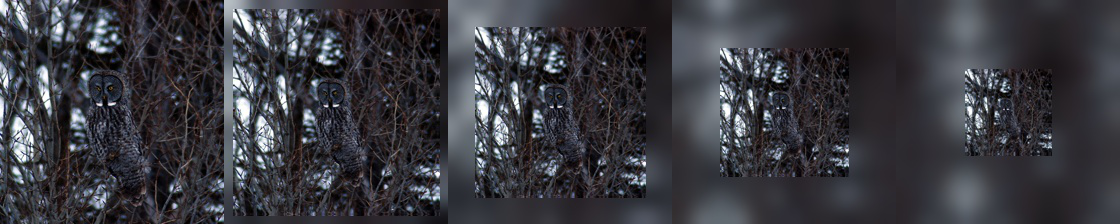}
         \caption{Re-scaled images from the ImageNet validation set}
         \label{fig:gradcam_vis_input}
     \end{subfigure}
     \hfill
     \begin{subfigure}[b]{0.45\textwidth}
         \centering
         \includegraphics[width=\textwidth]{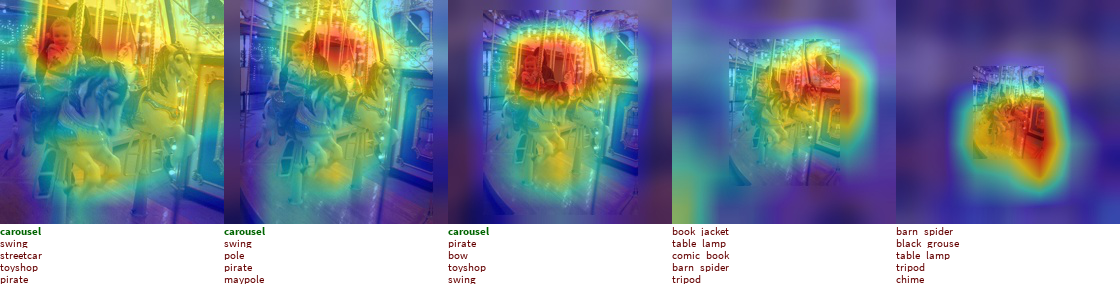}
         \includegraphics[width=\textwidth]{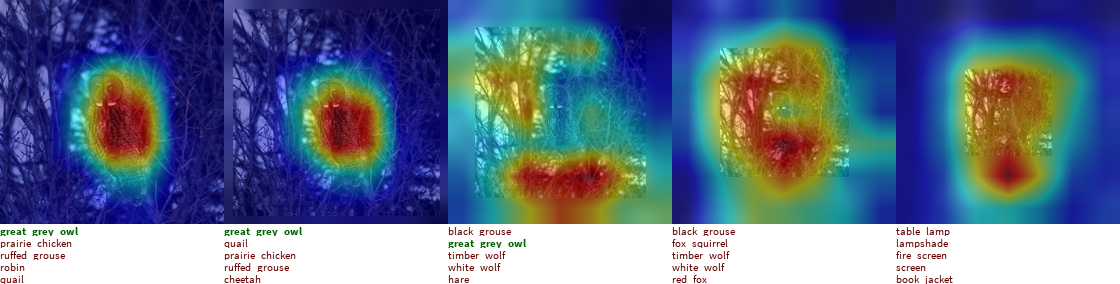}
         \caption{AutoAugment~\cite{AutoAugment}}
         \label{fig:gradcam_vis_AA}
     \end{subfigure}
     \hfill
     \begin{subfigure}[b]{0.45\textwidth}
         \centering
         \includegraphics[width=\textwidth]{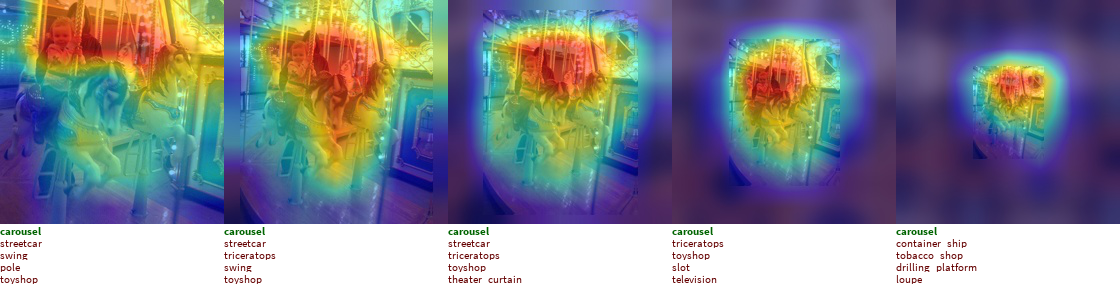}
         \includegraphics[width=\textwidth]{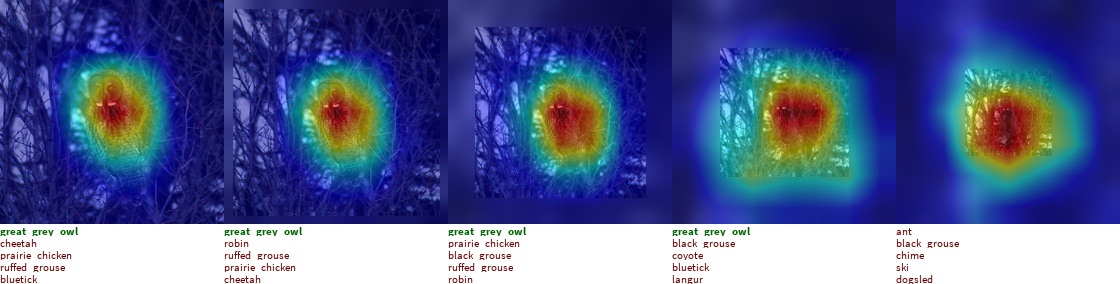}
         \caption{AutoAugment~\cite{AutoAugment} + InAug (ours)}
         \label{fig:gradcam_vis_InAug}
     \end{subfigure}
        \caption{Grad-CAM~\cite{GradCAM} visualization for ResNet50 (\textbf{best viewed when zoomed}). The input images appear in figure~\ref{fig:gradcam_vis_input}, where the leftmost column contains images that underwent the standard validation pre-processing. In figure~\ref{fig:gradcam_vis_AA} we show the Grad-CAM visualization for a network trained using AutoAugment. We also print the top-5 predictions of the network below each image. Labels are ordered by the network's confidence, and correct labels are highlighted in green. Similarly, we report the same visualization for ResNet50 trained with AutoAugment + InAugmet (ours) in figure~\ref{fig:gradcam_vis_InAug}.}
    \label{fig:gradcam_vis}
\end{figure}

\section{Experiments}
\begin{figure}
    \centering
    \includegraphics[width=\linewidth]{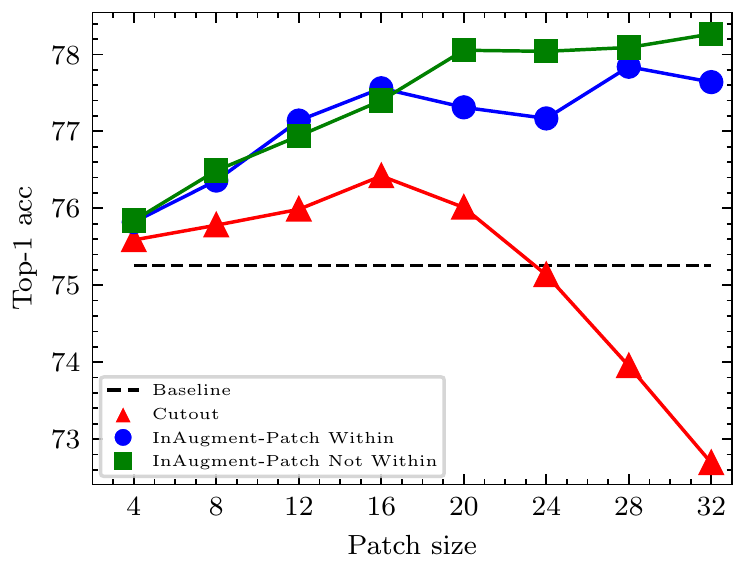}
    \caption{Comparison between CutOut~\cite{CutOut} and InAugment (ours). We plot PreAct-ResNet18~\cite{PreActResnet} top-one accuracy as a function of the patch size used for CutOut and InAugment. In all experiments, only the standard augmentations were applied on the base image (i.e., random crop and random horizontal flip). As can be seen from the figure, CutOut has a certain patch-size threshold, after which applying CutOut result in information loss. On the other hand, InAugment benefits from larger patches. For InAugment, we considered two setups. In the first setup, patches are restricted to be within the image boundary (blue), while in the other setup, patches can deviate from the image boundary (green). Note that the second setup yields better results, which indicates that varying patch-sizes could be beneficial. }
    \label{fig:inaugment_vs_cutout}
\end{figure}

In this section, we conduct several experiments to support our claim. First, we show a direct comparison between InAugment and Cutout~\cite{CutOut} and argue that InAugment achieves better results even for larger patch sizes. Second, we show that resizing the patches is beneficial and using multiple patches can boost the network's accuracy. We also show that networks trained with InAugment are robust to scale, especially for out-of-distribution object sizes. Finally, we compare InAugment with previous state-of-the-art augmentation techniques and evaluate the test-accuracy on two popular datasets, ImageNet ILSVRC12~\cite{ILSVRC12} and the CIFAR~\cite{CIFAR} dataset. Alternative approaches to InAugment are reported in the supplemental materials. 

\subsection{Ablation experiments}
\label{sec:InaugmentVsCutout}
\label{sec:resizing_effictevness}

In this section, all networks are trained on the CIFAR-100 dataset. Unless otherwise stated, then the training setup follows the one in Section~\ref{sec:cifar_results}.

\paragraph{InAugment vs CutOut~\cite{CutOut}:} CutOut is an augmentation method in which random regions of the input image are removed. The idea in CutOut, is that eliminating certain parts of the image could encourage the network to look at other features and essentially serve as a regularization technique. This technique is very effective on the CIFAR dataset, and it is usually adopted when training neural networks on this dataset. 

In our method, random patches are copied and pasted on the image, which could lead to the case where non-important regions (e.g., background patches) are pasted on top of essential areas. Therefore, InAugment may be performing CutOut, where important parts are removed continuously by pasting non-important regions on top of them. To see that this is not the case, we trained our network in three different settings. In one experiment, we apply CutOut augmentation with varying cut sizes. For InAugment, we only copy one random patch and randomly paste it back on the input image after augmentation (we do not perform any resize). We conduct two experiments for InAugment, one where the patch is entirely within the image boundary. In the second experiment, we allow patches to deviate from the image boundary. In this case, the patch is trimmed to only copy pixels from the image itself (we do not perform any padding). Finally, we also trained the network using the standard augmentation (i.e., random crop and random horizontal flip) and reported baseline accuracy.

In Figure~\ref{fig:inaugment_vs_cutout}, we report the average top-1 accuracy as observed over five different runs. As can be seen, both CutOut and InAugment improve over the baseline training. One of the main issues with the Cutout is that removing larger areas usually leads to loss of information, and the effective cut size for CIFAR100 is 16. Interestingly, with InAugment, larger patch sizes only improve the network's performance and lead to a ~2\% performance boost in network accuracy. One reason for this phenomenon is that it is less likely that these large patches will contain only useless information when copied from the image. Hence, the pasted patch will repeat essential features in the input images while still occluding other parts. Lastly, note that when we allow patch sizes to vary, we achieve a 0.4\% performance gain relative to constant size patches. Therefore, applying varying patch sizes could lead to better results. It is worth mentioning that this was strongly observed in the ImageNet~\cite{ILSVRC12} experiments, where the input images have varying patch sizes. In our experiments, we found that copying patches with random-sizes are essential in the ImageNet experiment.


\paragraph{Effect of resizing patches:} the repetition of patches in varying scales is an integral part of InAugment. To see the effectiveness of resizing the patches, we experiment with copying and pasting a single patch. In particular, we copy a random patch of size $s_p \times s_p$ and resize it to size $s_r \times s_r$, where $s_p, s_r \in \{ 12, 16, 20, 24\}$ (note, in the case where $s_p = s_r$, then we don't perform any resizing). In this experiment, both the input image and the patch undergo the standard transformation (e.g. random horizontal flip).

\begin{table}[h]
\centering
\resizebox{0.75\linewidth}{!}{%
\begin{tabular}{@{}cccccc@{}}
\multicolumn{1}{l}{}       & \multicolumn{1}{l}{} & \multicolumn{4}{c}{Patch resize ($s_r$)}            \\ \cmidrule(l){2-6} 
                           &                      & 12    & 16    & 20    & 24    \\ \cmidrule(l){2-6} 
\multirow{4}{*}{\rotatebox[origin=c]{90}{Patch size ($s_p$)}} & 12                   & 76.78 & 77.45 & 77.63 & \textbf{77.73} \\ \cmidrule(l){2-6} 
                           & 16                   & 77.65 & 77.45 & \textbf{78.01} & 77.66      \\ \cmidrule(l){2-6} 
                           & 20                   & 77.33 & 77.87 & 77.67  & \textbf{78.02}   \\ \cmidrule(l){2-6} 
                           & 24                   & \textbf{77.93} & 77.78 & 77.58 & \textbf{77.9}       \\ \cmidrule(l){2-6} 
\end{tabular}}
\vspace{5pt}
\caption{Resize affect on top-1 accuracy. Rows represent the size of the copied patch. Columns represent the new size of the patch before we paste it (after resizing). We also report the accuracy in the case we don't perform any resize (diagonal).}
\label{table:resize_effect}
\end{table}

As can be seen from Table~\ref{table:resize_effect}, the best accuracy obtained when we resize the patches before pasting them back into the image. In particular, the smaller the patch, the more crucial resizing become. For example, the best accuracy reported for patches of size $(12\times 12)$, is when we resize the patch to $(24 \times 24)$, which is $1 \%$ improvement over not performing any resize. For larger patches, we see that the resizing helps for most $s_r$. We believe this is because in our implementation, if the random patch we  try to copy deviates from the image boundary, then it is trimmed to fit the image boundary. Therefore, larger patches are frequently clipped, which essentially means that on average their effective size is small.

\paragraph{Effect of multiple patches: } To see the effect of pasting multiple patches onto the image, we trained three models, each with three different settings, added on top of the AutoAugment baseline. For settings-$k$, we copy $k$ random patches of size $(32 \times 32)$. The patches are then resized such that the $i$-th patch is re-scaled by factor $\sigma_i = (0.5)^{i-1}$.

\begin{table}[]
\centering
\resizebox{0.95\linewidth}{!}{%
\begin{tabular}{@{}lcccc@{}}
\toprule[1.5pt]
\multicolumn{1}{c}{}   & AA      & InAug x1 & InAug x2 & InAug x3 \\ \midrule
PreActResNet       & 79.11  / 0.95   & \textbf{80.27} / 0.82  & 80.19 / \textbf{0.79}         & 79.70 / 0.80  \\
WideResNet-28-10      & 83.84 / 0.59 & \textbf{84.80} / \textbf{0.54}  & 84.46 / \textbf{0.54}  & 84.27 / 0.55  \\
ShakeShake-96         & 85.90 / 0.55  & 86.65 / 0.52  & \textbf{86.89} / \textbf{0.50}  & 86.65 / \textbf{0.50}  \\ \bottomrule[1.5pt]
\end{tabular}
}
\vspace{5pt}
\caption{Number of patches effect. We report the average top-1 accuracy and the test Cross Entropy Loss as observed over 5 different runs. In this experiment we copy and paste up-to three patches. We let InAugment x$n$ represent the experiment in which $n$-patches are used (where $n=1,2,3$). We also report the results obtained for Auto Augment using our implementation.}
\label{table:number_of_patches}
\end{table}

The average test accuracy and the average test loss are reported in Table~\ref{table:number_of_patches} for all three settings. Also, we report the results obtained by training the models using AutoAugment as a baseline augmentation technique. As can be seen from the Table, incorporating InAugment not only improves the overall accuracy but significantly improves the Cross-Entropy (CE) loss on the test-set. Interestingly, training Pre-ActResNet18~\cite{PreActResnet} with two patches improves the average CE loss by 0.03, even though the average accuracy is somewhat negatively affected. This suggests that the network's confidence improves when increasing the number of patches. We believe the network accuracy has not improved because the network has a small capacity, and using multiple patches makes the optimization process harder.  On the other hand, training ShakeShake26-96 with InAugmentx2 is the best setting for that network. We conclude that using multiple-patches becomes more significant when training larger models for a longer period of time. 

\subsection{CIFAR Experiments}
\label{sec:cifar_results}
\begin{table*}[h]
\centering
\resizebox{0.8\textwidth}{!}{%
\begin{tabular}{@{}lccccccc@{}}
\toprule[1.5pt]
                 & \multicolumn{1}{l}{\textbf{Baseline}} & \multicolumn{1}{l}{\textbf{CutOut}} & \textbf{AA}  & \multicolumn{1}{l}{\textbf{PBA}} & \multicolumn{1}{l}{\textbf{FAA}} & \multicolumn{1}{l}{\textbf{RA}} & \multicolumn{1}{l}{\textbf{AA+InAug}} \\ \midrule
\textbf{CIFAR10} \\
PreAct-ResNet-18 & 94.32 $\pm$ .18                          & 95.67 $\pm$ .15                        & 96.0 $\pm$ .05 & -                                & -                                & -                               & \textbf{96.35 $\pm$ .08}                  \\
WideResNet-28-10 & 96.1                                  & 96.9                                & 97.4         & 97.4                             & 97.3                             & 97.3                            & \textbf{97.45 $\pm$ .04}                  \\
ShakeShake26\ 2x96 & 97.1                                  & 97.4                                & 98.0         & 98.0                             & 98.0                             & 98.0                            & \textbf{98.30 $\pm$ .05}                  \\ \midrule

\textbf{CIFAR100} \\
PreAct-ResNet-18 & 75.32 $\pm$ 0.11                          & 76.48 $\pm$ 28                        & 79.11 $\pm$ .11 & -                                & -                                & -                               & \textbf{80.27 $\pm$ .31}                  \\
WideResNet-28-10 & 81.2                                  & 81.6                                & 83.80 $\pm$ .17         & 83.3                             & 82.7                             & 83.3                            & \textbf{84.80 $\pm$ .20}                  \\
ShakeShake26\ 2x96 & 82.9                                  & 84.0                                & 85.90 $\pm$ .11         & 84.7                             & 85.4                             & -                            & \textbf{86.89 $\pm$ .20}  

                                                                                                                    \\ \bottomrule[1.5pt]
\end{tabular}%
}
\vspace{5pt}
\caption{Top-1 accuracy report on CIFAR. The reported results with both mean and std are based on our implementation. Results for AutoAugment (AA), Population-Based Method (PBA) Fast AutoAugment (FAA) and RandAumgnet (RA) are reported in the original papers of these methods~\cite{RandAugment,AutoAugment, PBA, FastAutoAugment}. In particular, for AA we report the better result between our implementation and those reported in the original paper~\cite{AutoAugment}. }
\label{table:cifar100_results}
\end{table*}
We compare InAugment with previous augmentation methods: CutOut~\cite{CutOut}, AutoAugment (AA)~\cite{AutoAugment}, Population Based Augmentation (PBA)~\cite{PBA}, FastAutoAugment (FAA)~\cite{FastAutoAugment}, and RandAugment (RA)~\cite{RandAugment}. In our experiment, we train three different networks, PreAct-ResNet18~\cite{PreActResnet, ResNet}, WideResNet-28-10~\cite{WideResNet}, and Shake-Shake26-2x96~\cite{ShakeShake} on the CIFAR10 and CIFAR100 datasets~\cite{CIFAR}. 

\textbf{Experiment details: } for all experiments we use the Stochastic Gradient Descent (SGD) optimizer, with momentum 0.9 and weight decay of $\tt{1e-3}$, $\tt{1e-4}$, and $\tt{5e-4}$ for PreAct-ResNet18, WideResNet28-10 and ShakeShake26-2x96, respectively. A nesterov momentum we used in the experiments of WideResNet28-10 and Shake-Shake26-2x96. The training settings for PreAct-ResNet18 matches those in~\cite{Mixup}, and for WideResNet and Shake-Shake26-2x96, we follow the training settings used in~\cite{AutoAugment, RandAugment}. In particular,  we train PreAct-ResNet18 and WideResNet-28-10 for 200 epochs with the initial learning rate set to 0.1. For Shake-Shake26-2x96, we use a learning rate 0.01 and train the network for 1800 epochs. The learning rate schedule used for PreAct-ResNet18 is a multi-step scheduler, in which the learning rate is scaled by 0.1 on epochs 100 and 150. A cosine-learning~\cite{CosineLR} rate was used for both WideResNet-28-10 and Shake-Shake26-2x96. For the WideResNet28-10 and PreAct-ResNet18 models, we copy one patch of size $(32 \times 32)$ and don't perform any resize on the patch. For Shake-Shake26-2x96 model, we copy two patches of size $32$, and rescale the second patch by $0.5$. Also, we always chose to paste the copied pasted, i.e., our drop rate is $p_i=1$ (see section~\ref{sec:paste_stage}).

\textbf{Results:}  the results are shown in Table~\ref{table:cifar100_results}. Specifically, incorporating InAugment with AutoAugment yielded the highest accuracy for all networks used. We obtain about 1.0\% improvement for all networks trained on the CIFAR100 dataset compared to the best-reported result for all the other methods. For CIFAR10, we also achieve the highest accuracy compared to all the networks and augmentation methods on which we experimented, although by a smaller margin. We postulate the reduced improvement is due to the high initial accuracy, and the heavily tuned regularization schemes used by some of the methods. We note that our experiments were impelemented using the Pytorch framework, while the AA baseline was originally implemented using the Tensorflow one. This induces minor changes in the resulting accuracies ( e.g., in our implementation, the WideResNet28-10 model trained with AA achieves only 97.15\% as opposed to the 97.4\% reported in the original paper). In any case, we report in Table 3 the accuracy which is better between our implementation and that of the original paper~\cite{AutoAugment}. 


\subsection{ImageNet Experiments}
We also show that InAugment is effective on larger datasets with high-resolution images. In particular, we train ResNet-50~\cite{ResNet} and EfficientNet-B3~\cite{CIFAR} from scratch on the ImageNet-ILSVRC2012~\cite{ILSVRC12} dataset.

\textbf{Experiment details: } the training setup follows the default implementation in the official TensorFlow-1.15.4 code. In particular, the ResNet50 models were trained for 180 epochs and the EfficientNet-B3 models were trained for 350 epochs.  We use the same training hyper-parameters used in~\cite{RandAugment, AutoAugment}. The training images follow the standard augmentation where they are randomly-cropped and resized to the size $(224 \times 224)$ for the ResNet50 model, and to size $(300 \times 300)$ for the EfficientNet-B3 model (a random horizontal flip is applied). During test-time, images are center cropped and resized to the same resolution as in the training. The ResNet50 network was trained with the SGD optimizer, with momentum 0.9 and weight decay $\tt{1e-4}$. We trained the network on a single v3-8 TPU, with a global batch size of 1024. The learning rate was set to 0.1 and is linearly scaled by the batch-size divided by 256, following~\cite{LearnInOneHour}. To train EfficientNet-B3, we use a single v2-28 TPU pod with global batch size of 4096. The network was trained with RMSProp with learning rate of 0.016, momentum 0.9, $\epsilon=0.001$ and decay of 0.9. The learning rate is also linearly scaled as in the case of the ResNet50 model.

The base augmentation we use is the policy that AutoAugment found for Efficientnet~\cite{EfficentNet} (this policy is available in the official TensorFlow code). Furthermore, we adopt the resize-first implementation (see copy-stage in subsection~\ref{sec:copy_stage}) since it achieves the right balance between efficiency and accuracy. Specifically, the pre-processing time is lower when applied to a low-resolution image; therefore, augmenting the resized patches will take less time (which is crucial for large-scale datasets). For ResNet50, we copy three patches of random sizes and resize them to $(134 \times 134)$, $(80 \times 80)$, and $(48 \times 48)$, the patches are then pasted according to the order of their size. For Efficientnet-B3~\cite{EfficentNet}, we copy two-patches and randomly resize them to size $(s_1 \times s_1)$ and  $(s_2 \times s_2)$, where $s_1$  and $s_2$ are uniformly sampled from $[300, 150]$ and $[150, 8]$, respectively. In both experiments, patches are dropped and not pasted onto the images with a probability 0.5.

\begin{table*}[]
\centering
\resizebox{0.8\linewidth}{!}{%
\begin{tabular}{@{}lccccc@{}}
\toprule[1.5pt]
                & Baseline    & Fast AA     & RA          & AA          & AA + InAug (Ours) \\ \midrule[1.5pt]
ResNet-50       & 76.3 / 93.1 & 77.6 / 93.7 & 77.6 / 93.8 & 77.6 / 93.8 & \textbf{78.2} / \textbf{94.0 }      \\
EfficientNet-B3 & 81.1 / -    & -           & -           & 81.6 / -    & \textbf{81.8} / 95.6       \\ \bottomrule[1.5pt]
\end{tabular}
}
\vspace{5pt}
\caption{Top-1 and Top-5 accuracy results on ImageNet. The results for AutoAugment on ResNet50 are replicated in our experiments. The results for other methods and models are from their original papers~\cite{RandAugment, FastAutoAugment, AutoAugment}}

\label{table:imagenet_results}
\end{table*}
\textbf{Results:} the results are reported in Table~\ref{table:imagenet_results}. As shown from the table, we improve both the top-1 and top-5 accuracy of ResNet-50 compared with previous augmentation methods. Specifically, we achieve 78.2-\% top-1 accuracy, which is about $0.6\%$ improvement over previous augmentation methods~\cite{AutoAugment, FastAutoAugment, RandAugment}. We also exhibit an approximately $0.2 \%$ improvement in the top-5 accuracy. Similarly, we also witnessed an improvement in the Efficientnet-B3 experiments, where a top-1 accuracy of $81.8\%$ was achieved, which is $0.2\%$ higher than using AutoAugment as the only augmentation.

\subsection{Out-of-distribution samples} 

The standard ImageNet~\cite{ILSVRC12} training consists of randomly cropping images and resizing them to a fixed size (e.g., $224 \times 224$ for ResNet50~\cite{ResNet}). Therefore, the network will most likely observe the same object at different sizes during training. However, we show that this pre-processing is not enough to handle out-of-distribution scales, and using InAugment in training achieves a robust model that is more scale-invariant.  

\begin{figure}
\centering
\includegraphics[width=0.95\linewidth]{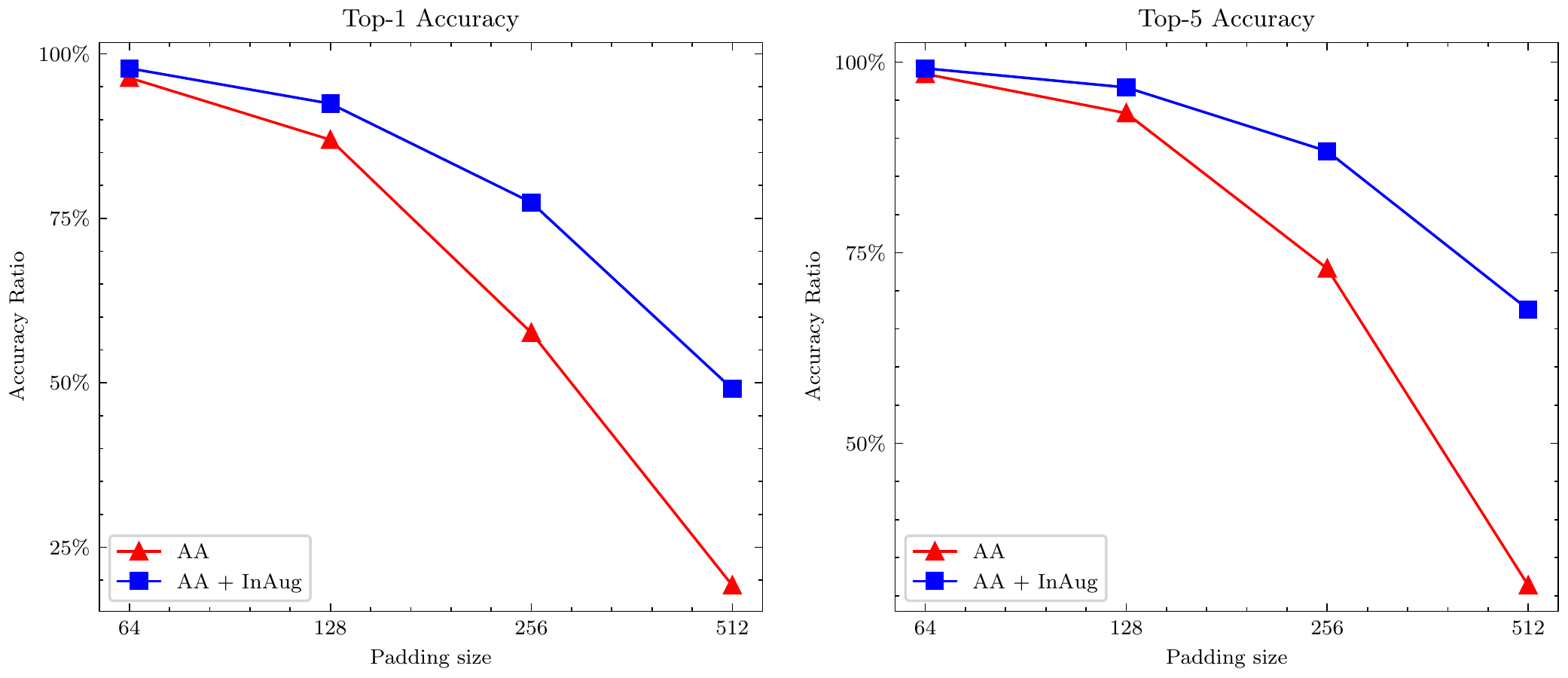}
\caption{ResNet50~\cite{ResNet} robustness to scale. We plot the ratio between the network's accuracy after scaling the validation set and its accuracy on the original validation set. The ratio of both the top-1 (left) and the top-5 accuracy (right) are shown. As can be seen, training the network using AutoAugment (AA) + InAugment (InAug) achieves significantly better results than training it solely with AA augmentation. }
\label{fig:ood_comparison}
\end{figure}

To test our hypothesis, we want to measure the network's ability to classify objects at different scales. To scale the entire ImageNet's validation set, we pad each image with fixed padding $D$, and the rest of the pre-processing remains the same.  Furthermore, to avoid introducing artifacts, we use symmetric padding and strongly blur the padded pixels. Note that we cannot simply resize the images since ResNet50 was designed to process images of size at least $224 \times 224$. We create four different validation sets, using different pad sizes, namely $D \in \{64, 128, 256, 512 \}$ (see figure~\ref{fig:gradcam_vis_input}).

In figure~\ref{fig:ood_comparison} we plot the ratio of the network's accuracy on the padded validation sets and the original set. As can be seen, incorporating InAugment in pre-processing the training data yields models with better robustness to scale. In particular, you can see that for large padding (i.e., smaller scales), the network's performance trained solely using Auto Augment witnesses a catastrophic performance drop. Note that performance drop on both settings is inevitable since some images already appear on a small scale (e.g., in the owl image in figure~\ref{fig:gradcam_vis_input}, you can see that it is difficult to distinguish the owl in the small scale example). We also visualize the regions that the network bases its decision on using GradCAM~\cite{GradCAM}. In figure~\ref{fig:gradcam_vis_AA}, it is evident that the baseline network no-longer looks at important regions when it makes its decision on small-scale images. On the other hand, as seen in figure~\ref{fig:gradcam_vis_InAug}, models trained with our methods exhibit localized decisions for smaller-scale instances. Note in the owl example, even though the network trained with our method did not correctly classify the image with padding $D=512$, it still based its decision on the owl location. However, due to the extreme scale, the owl features are no longer distinguishable. 

Finally, we also considered two alternative approaches to scale the validation set. In one setup, we used zero-padding, and in the other, we tiled the image multiple times. In both cases, training with our-method improves upon solely training with Auto Augment. Especially in the zero-padding case, where the network trained with Auto Augment only, confuses the padded-images to be 'Television' (see supplemental material for more examples).

\section{Discussion}
\label{sec:discussion}
In this paper, we introduced a novel image augmentation method. We showed that copying-and-pasting random patches expose the network to higher quality features, especially when the patches are resized. In particular, InAugment improves the robustness of the network to scale and improves its confidence and accuracy. As mentioned throughout the text, we compared InAugment with previous augmentation methods, including AutoAugment~\cite{AutoAugment}, RanAugment~\cite{RandAugment}, Fast AutoAugment~\cite{FastAutoAugment}, and CutOut~\cite{CutOut}, showing consistent improvement. We believe that incorporating InAugment with existing methods could further boost classifiers' performance. For example, InAugment can be considered an image transformation function, which means it can also be added to automated augmentations' search space (e.g., AutoAugment). Furthermore, we believe that performing different augmentations on the copied patches could increase the image's information variance, boosting the network's performance even further. Beyond improving validation accuracy on the test set, some augmentation techniques corrupt~\cite{Robustness, Robustness1, Robustness2, Robustness3} the input data and lead to robust models. The same distortions could also be applied to the copied patches, which could increase the distortion variance. Therefore, we believe InAugment can also be adapted for improving the model robustness for out-of-distribution images.  Finally, since patches are copied randomly in InAugment, it might be fruitful to investigate attention-based InAugment, where meaningful patches are copied instead of random ones and are pasted onto non-important regions.
\section*{Acknowledgment}
Research supported with Cloud TPUs from Google's TensorFlow Research Cloud (TFRC).

{\small
\bibliographystyle{ieee_fullname}
\bibliography{egbib}
}

\end{document}